\documentclass{lib/bmvc2k}

\title{Quantitative Metrics for Evaluating Explanations of Video DeepFake Detectors}

\addauthor{Federico Baldassarre}{fedbal@kth.se}{1,\(\ast\)}
\addauthor{Quentin Debard}{quentin.debard@huawei.com}{2}
\addauthor{Gonzalo Fiz Pontiveros}{gonzalo.fiz.pontiveros@huawei.com}{2}
\addauthor{Tri Kurniawan Wijaya}{tri.kurniawan.wijaya@huawei.com}{2}

\addinstitution{
 KTH - Royal Institute of Technology\\
 Stockholm, Sweden
}
\addinstitution{
 Huawei Ireland Research Center\\
 Georges Court, Townsend St,\\
 Dublin, Ireland
}

\runninghead{Baldassarre et al.}{Quantitative Metrics for Evaluating Explanations}

\def\wrt{\emph{w.r.t}\bmvaOneDot}
\def\ie{\emph{i.e}\bmvaOneDot}
\def\eg{\emph{e.g}\bmvaOneDot}

\def\etal{\emph{et al}\bmvaOneDot}

\usepackage{graphicx}
\usepackage{subcaption}
\usepackage{amsmath}
\usepackage{amssymb}
\usepackage{booktabs}

\usepackage{mathtools}
\usepackage{bm}
\usepackage{amsthm}
\usepackage{verbatim}
\usepackage{blindtext}
\usepackage{placeins}
\usepackage[inline]{enumitem} %
\usepackage[acronym]{glossaries}

\makeatletter
\newcommand*{\glsplainhyperlink}[2]{%
  \colorlet{currenttext}{.}%
  \colorlet{currentlink}{\@linkcolor}%
  \hypersetup{linkcolor=currenttext}%
  \hyperlink{#1}{#2}%
  \hypersetup{linkcolor=currentlink}%
}
\let\@glslink\glsplainhyperlink
\makeatother

\usepackage[capitalize]{cleveref}
\crefname{section}{Sec.}{Secs.}
\Crefname{section}{Section}{Sections}
\crefname{table}{Tab.}{Tabs.}
\Crefname{table}{Table}{Tables}

\newcommand*{\sota}{state-of-the-art\@\xspace}
\newcommand*{\df}{DeepFake\@\xspace}
\newcommand*{\dfs}{DeepFakes\@\xspace}
\newcommand*{\exn}[2]{\mathbb{E}_{#2}\left [#1 \right ]}
\newcommand*{\dfdc}{DeepFake Detection Challenge\@\xspace}
\newcommand*{\dfd}{DeepFake Dataset\@\xspace}

\DeclarePairedDelimiter{\norm}{\lVert}{\rVert}

\newacronym{DL}{DL}{Deep Learning}
\newacronym{XAI}{XAI}{Explainable AI}
\newacronym{GAN}{GAN}{Generative Adversarial Network}
\newacronym{DFDC}{DFDC}{DeepFake Detection Challenge}
\newacronym{DFD}{DFD}{DeepFake Detection Dataset}
\newacronym{TV}{TV}{Total Variation}

\definecolor{bmvcblue}{rgb}{0.0, 0.0, 0.4}

\newcommand\blfootnote[1]{%
  \begingroup
  \renewcommand\thefootnote{}\footnote{#1}%
  \addtocounter{footnote}{-1}%
  \endgroup
}

\begin{document}

\maketitle
\blfootnote{\textsuperscript{*} Work performed during an internship at Huawei Ireland Research Center}
\begin{abstract}
The proliferation of \df technology is a rising challenge in today's society, owing to more powerful and accessible generation methods.
To counter this, the research community has developed detectors of ever-increasing accuracy.
However, the ability to explain the decisions of such models to users is lacking behind and is considered an accessory in large-scale benchmarks, despite being a crucial requirement for the correct deployment of automated tools for content moderation.
We attribute the issue to the reliance on qualitative comparisons and the lack of established metrics.
We describe a simple set of metrics to evaluate the visual quality and informativeness of explanations of video \df classifiers from a human-centric perspective.
With these metrics, we compare common approaches to improve explanation quality and discuss their effect on both classification and explanation performance on the recent DFDC and DFD datasets.
\end{abstract}

\section{Introduction}
\label{sec:intro}

``\df'' refers to the realistic alteration or generation of multimedia content, in visual, audio, or textual form. 
The most striking application of \dfs are generative deep learning models that can alter a person's appearance in videos.
From early attempts~\cite{thies_face2face_2016,suwajanakorn_synthesizing_2017}, the quality of these \emph{face-swapping} techniques has increased consistently to the point that both casual and attentive observers can be fooled.
While some applications can be positively innovating~\cite{westerlund_emergence_2019}, \dfs can be designed with malicious intent, such as online disinformation or public defamation.
In response, the research community has introduced datasets~\cite{rossler_faceforensics_2019,nickdufour_dfd_2019,dolhansky_deepfake_2020,pu_deepfake_2021} and methods~\cite{coccomini_combining_2021,bonettini_video_2021,kim_fretal_2021} for the automatic monitoring and detection of \dfs.
However, benchmark performance has become the \textit{de-facto} goal, shadowing other aspects that are crucial for the correct deployment of such models.

In practice, as automated \df detectors acquire a significant role for moderation and censorship of online communities, it becomes necessary to inspect and \emph{explain} their decision process.
From the users' perspective, it is not acceptable that ``black-box'' models manage their freedom of expression and online safety.
Instead, users require intuitive explanations to validate \df forgeries, prevent unjustified censorship, and trust automated moderation systems. 
From the perspective of companies and regulators, interpretability is necessary to justify the enforcement of \df detectors, in accordance to the \emph{right to explanation} of legal frameworks such as the GDPR~\cite{europeanunion_gdpr_2018}.
Also, developers of such tools can benefit from explanations to verify the learned representation, mitigate unwanted bias, and defend against adversarial attacks.

\begin{figure}[t]
  \centering
  \includegraphics[width=.45\linewidth]{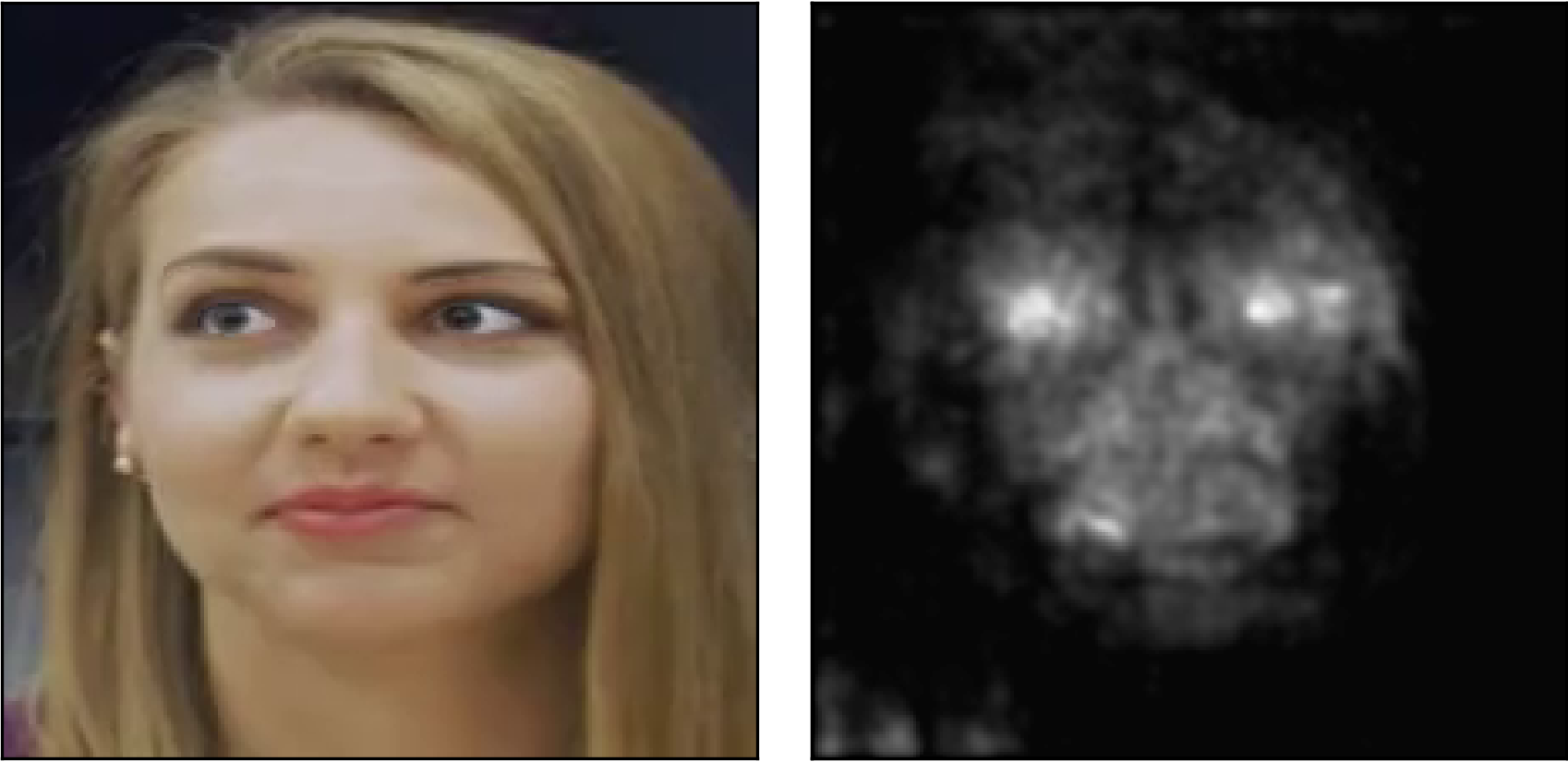}
  \caption{\textbf{Explanation heatmap} for one sample from \dfd obtained by applying SmoothGrad to a classifier regularized with Total Variation. The resulting heatmap is visually smooth (\(\text{TV}=0.25\)), localized (\(\sigma=731\)), and concentrates around visible manipulation artefacts (\(M_\text{in}=18.1\%\)), \ie the eyes and one corner of the mouth.}
  \label{fig:teaser}
\end{figure}

Several methods for explaining visual classifiers exist, \eg~\cite{simonyan_deep_2014,selvaraju_gradcam_2020}, which can be compared in terms of faithfulness to the model and correctness to the data.
However, researchers lack quantitative tools to evaluate human-centric properties of explanations and claims of improved informativeness are often based on subjective comparisons.
This work introduces a quantitative framework to evaluate \df explanations \wrt to human perception, which can be applied in practical deployments of \df classifiers.
In particular, we contextualize existing metrics, \ie manipulation detection, and propose new ones as needed, namely for \emph{smoothness}, \emph{sparsity}, and \emph{locality}.
We apply these metrics to \sota video recognition models and compare several techniques intended for improving explanations, forming a quantitative baseline on the \dfdc dataset and the \dfd~\cite{rossler_faceforensics_2019,dolhansky_deepfake_2020}.
Last, we empirically evaluate how to best communicate heatmap-based explanations to users, discuss limitations and future directions for \df explainability.

\section{Related work}
\label{sec:related_work}

\paragraph{\df generation.}
Since their inception, generative models have been applied to manipulate faces, bodies and voices in online media.
Today's availability of online content and ease of access to open-source frameworks, allow anyone with consumer-grade hardware to generate \dfs.
While legitimate applications of this technology exits, \eg dubbing, \dfs have been infamously used for disinformation, fraud, hatred, sexual abuse, and other crimes~\cite{damiani_voice_2019,vaccari_deepfakes_2020}.
This work focuses on visual forgery of faces in videos, which can be categorized as \emph{face swapping}, in which the appearance of a face is replaced with another~\cite{nirkin_face_2018,deepfakesteam_deepfakes_2021,kowalski_faceswap_2021,li_advancing_2020}; or \emph{facial reenactment}, in which expressions are edited~\cite{thies_face2face_2016,thies_deferred_2019}.
Such manipulations can be produced via purely learning-based generative models \cite{nirkin_fsgan_2019,deepfakesteam_deepfakes_2021,thies_deferred_2019} or hybrid computer graphics approaches \cite{thies_face2face_2016,kowalski_faceswap_2021}.
For survey of methods and applications, we refer the reader to the works of Tolosana~\etal\cite{tolosana_deepfakes_2020} and Masood~\etal~\cite{masood_deepfakes_2021}.

\paragraph{\df detection.}
In response to the widespread misuse of \dfs, researchers and companies have started focusing on automated detection of forged media.
Forensic approaches vary from detecting anatomical inconsistencies~\cite{agarwal_protecting_2019,li_ictu_2018,li_exposing_2018}, to analyzing digital artefacts~\cite{afchar_mesonet_2018,yang_exposing_2019,matern_exploiting_2019}.
Other approaches are purely learning-based~\cite{guera_deepfake_2018,amerini_deepfake_2019} and can integrate advanced architectures and optimization techniques~\cite{nguyen_capsuleforensics_2019,nguyen_multitask_2019,bondi_training_2020,kim_fretal_2021,coccomini_combining_2021}.
Key to this effort is the release of large-scale datasets of images~\cite{karras_stylebased_2019a,dang_detection_2020,neves_ganprintr_2020,generatedphotosteam_generated_2018}, audio~\cite{zhou_talking_2019}, and videos~\cite{rossler_faceforensics_2019,nickdufour_dfd_2019,li_celebdf_2020,dolhansky_deepfake_2020,pu_deepfake_2021,zi_wilddeepfake_2021}, which allow to train deep models for forgery detection.

\paragraph{Explainable AI.}
Although powerful, deep learning models are often deemed ``black-boxes'' to illustrate the opacity of their decision process.
The field of study of Explainable AI (XAI) tries to address these shortcomings to allow users, researchers, and regulators to gain insights into such models (\emph{model interpretability}) and their outputs (\emph{decision explainability})~\cite{gilpin_explaining_2018,miller_explanation_2019}.
In the visual domain, in particular for classification, it is common to explain the decision of a model using heatmaps which highlight important areas of the input~\cite{zeiler_visualizing_2014, fong_interpretable_2017,petsiuk_rise_2018}.
Backpropagation-based approaches generate heatmaps by computing gradients~\cite{simonyan_deep_2014,zhou_learning_2015,smilkov_smoothgrad_2017,sundararajan_axiomatic_2017,kapishnikov_xrai_2019,selvaraju_gradcam_2020,xu_attribution_2020} or gradient surrogates~\cite{springenberg_striving_2015a,montavon_methods_2018,samek_explaining_2021}.
Alternative approaches construct proxy models that are locally faithful and easier to interpret, \eg LIME~\cite{ribeiro_why_2016}.
Recently, transformer models~\cite{vaswani_attention_2017,dosovitskiy_image_2020,arnab_vivit_2021} have popularized using attention maps as explanations~\cite{abnar_quantifying_2020,chefer_generic_2021,xu_show_2015}, although these might not be representative~\cite{jain_attention_2019}.

\paragraph{\df explainability.}
As social platforms integrate automated tools for \df detection and moderation in their pipelines~\cite{masood_deepfakes_2021}, it becomes crucial to offer proper justification when some content is blocked.
Prototype-based explanations as in Trinh~\etal~\cite{trinh_interpretable_2021}, could teach users to identify manipulation artefacts on their own.
Similarly, SHAP-based methods can be adapted to to videos by defining 3D super-pixels~\cite{lundberg_unified_2017,pino_what_2021}.
Focusing on input features, Wang~\etal~\cite{wang_humanunderstandable_2021} suggest pre-processing steps that result in more human-interpretable heatmaps, according to a qualitative evaluation.
Finally, human-annotated explanations, \eg Mathew \etal~\cite{mathew_hatexplain_2020}, provide direct insight on manipulation techniques.

\section{Method}
\label{sec:method}

\subsection{Explanations methods}
\label{sec:method-explanations}

Our goal is to establish quantitative metrics to evaluate explanations of visual \df classifiers.
In particular, we focus on heatmap-based methods~\cite{simonyan_deep_2014,zeiler_visualizing_2014,bach_pixelwise_2015} that associate each pixel to a scalar proportionally to its importance \wrt the classifier decision.
Formally, we define a video \(v\in\mathcal{V}\) as a mapping from a discrete grid \(\mathcal{G}= T\!\times\!H\!\times\! W\) to the RGB color space.
A \df classifier is then a function \(f: \mathcal{V} \rightarrow [0, 1]\) that maps a video to the probability distribution \(p(\textsc{fake}|v)\).
An explanation method is a function \(\Phi: \mathcal{V} \times \mathcal{F} \to \mathcal{H}\) that maps a pair \((v,f)\) to a relevance \emph{heatmap}  \(h:\mathcal{G} \rightarrow \mathbb{R}^+\), where \(\mathcal{F}\) and \(\mathcal{H}=\{h| \int h d\lambda = 1\}\) denote the set of classifiers and heatmaps respectively.
With this notation, popular gradient-based explanation methods are expressed as:
Sensitivity
\(\nabla f(v)\)~\cite{simonyan_deep_2014};
Gradient$\times$Input
\(\nabla f(v) \cdot v\)~\cite{kindermans_investigating_2016};
SmoothGrad
\(\mathbb{E}_{\epsilon\sim\mathcal{N}(0, \delta I)} \left[\nabla f(v + v_\epsilon)\right]\),
where $v_\epsilon$ adds random color perturbations~\cite{smilkov_smoothgrad_2017}; and 
Integrated Gradients
\((v-v_b)\cdot\int_0^1 \nabla f(v_b + \alpha(v-v_b)) d\alpha\),
where the baseline $v_b$ is a uniform black video~\cite{sundararajan_axiomatic_2017}.
Note that $\nabla$ and $\int$ are discretized operators over \(\mathcal{G}\) (see \Cref{app:explanation-metrics}).

Explanation methods are commonly compared according to their \emph{faithfulness}, \ie the ability to correctly explain a decision~\cite{bach_pixelwise_2015,ribeiro_why_2016,montavon_explaining_2017,alvarezmelis_robust_2018,selvaraju_gradcam_2020}.
Faithfulness is quantified by the \emph{deletion score}~\cite{samek_evaluating_2017,petsiuk_rise_2018} defined as
\(\exn{ \int f(v \odot (1 - h_\alpha)) d\alpha }{v}\), where \(h_\alpha\) is a binary mask obtained by selecting the most important pixels from the explanation \(\Phi(f, v)\) such that their cumulative relevance is \(\alpha\in[0, 1]\).
A low deletion score indicates a faithful explanation method: if relevant pixels are masked out first, the prediction confidence should drop sharply.
For our baseline model, SmoothGrad achieves the lowest deletion score (paired one-sided t-test \(p<10^{-5}\)) and is therefore selected for all evaluations of visual quality. We report per-method hyperparameters and per-dataset scores in \Cref{app:deletion-scores}.
Clearly, faithfulness is a necessary property of explanation methods, however, their heatmaps can still appear noisy and uninformative for humans, hence the need for quantitative metrics of visual quality.

\subsection{Evaluation metrics}
\label{sec:method-metrics}
As discussed in \Cref{sec:related_work}, several works address the representativeness or visual appearance of heatmaps.
However, the improvement is often demonstrated through qualitative examples, while quantitative comparison is lacking.
Understandably, defining general-purpose metrics for quantifying explanation properties is not trivial~\cite{ancona_better_2018}, as the perceived quality depends on the data itself, on the target user, and the downstream task.
Focussing on explanations of video \df classifiers, we discuss a set of desirable human-centric properties~\cite{koffka_principles_2013,barredoarrieta_explainable_2020,mohseni_multidisciplinary_2021} and formulate quantitative metrics for their evaluation.

\subsubsection{Visual quality}
\label{sec:method-metrics-visual-quality}
The first set of metrics considers general properties of explanation heatmaps that facilitate their understanding and communicability.
Complex models can take decisions based on features that are not easily accessible to users, \eg texture details or high-frequency patterns~\cite{geirhos_imagenettrained_2018,wang_humanunderstandable_2021}.
Instead, we expect models that focus on \emph{human-interpretable cues} \cite{koffka_principles_2013} such as small manipulation artefacts, teeth misalignment, non-circular pupils, or irregular skin complexion, to produce \emph{smoother}, \emph{sparser} and more \emph{localized} heatmaps.

\paragraph{Smoothness.}
Explanations that vary excessively between neighboring pixels or frames are not meaningful to humans~\cite{wang_humanunderstandable_2021}.
The smoothness of a heatmap  \(h:\mathcal{G} \to \mathbb{R}^{+}\) is measured as its \acrfull{TV}, where low values indicate higher local consistency:
\begin{equation}
\label{TV}
    \text{TV}(h) = \int_{\mathcal{G}} \norm{\nabla h}_1 d\lambda.
\end{equation}

\paragraph{Spatial locality.}
Unambiguous explanations should concentrate on few spatially-close patches of a video, \ie their relevance should be localized.
If we consider \(h\) as the distribution of a random vector \(\bm{\rho} \in \mathcal{G}\), we can measure locality through the volume of its variance matrix:
\begin{equation}\label{eq:det-covariance}
    \sigma=|\det(\bm{\Sigma})| = 
    \left| \det \left(
        \exn{\bm{\rho} \bm{\rho}^T}{h} - \exn{\bm{\rho}}{h}\exn{\bm{\rho}^T}{h}
    \right) \right|.
\end{equation}
A low \(\sigma\) will favor sharp unimodal distributions, \eg a Gaussian with low dispersion, as opposed to scattered multimodal heatmaps.
In the context of \dfs, this means highlighting single manipulation artefacts instead of allocating mass to distinct parts of the face.
For other tasks, spatial locality can be extended to account for domain-specific requirements.

\paragraph{Sparsity.}
While \acrshort{TV} and \(\sigma\) capture spatial properties, the individual values shall also be sparse, since few highly important regions are more indicative of a good explanation than several mildly relevant ones.
Both \(L_0\) norm and Entropy~\cite{shannon_mathematical_1948} are popular measures of sparsity, but the Gini Index~\cite{gini_variabilita_1912} is preferred according to Hurley and Rickard~\cite{hurley_sparsity_2009}.
For a heatmap \(h:\mathcal{G} \to \mathbb{R}^+\) and sorting indices \(i=\{1,\ldots,THW\}\) such that \(h(\bm{\rho}_i) \leq h(\bm{\rho}_{i+1})\):
\begin{equation}
    \operatorname{G} = \frac{2}{THW} \frac{\sum_i i \cdot h(\bm{\rho}_i)}{\sum_i h(\bm{\rho}_i)} - \frac{THW+1}{THW}.
\end{equation}

\subsubsection{Manipulation detection}
\label{sec:method-metrics-manipulation-detection}
Smooth, sparse and localized heatmaps appear visually appealing, but do they convey the location of manipulation cues?
Offering specific evidence greatly increases trust in the model, helps diagnosing failure cases, and encourages users to develop a critical eye for spotting \dfs.
In the \acrshort{XAI} literature, manipulation detection is commonly evaluated through user studies~\cite{selvaraju_gradcam_2020,wang_humanunderstandable_2021}, which suffers from reproducibility issues, or under a weakly-supervised paradigm~\cite{cao_look_2015,kolesnikov_seed_2016,baldassarre_explanationbased_2020,selvaraju_gradcam_2020}, which risk introducing bias from the additional annotations.

We argue that \dfs offer a unique possibility for the objective evaluation of weakly-supervised \emph{manipulation detection}.
Given a real video \(v_R\), its fake(s) \(v_F\), and a face parsing model \(s: \mathcal{G}\rightarrow \mathcal{P}\) that maps pixels of \(v_R\) to \(\mathcal{P} = \{\text{eyes}, \text{nose}, \text{mouth}\}\), an \textit{ad-hoc} evaluation sample can be produced such that the manipulation is limited to a specific semantic region:
\begin{equation}
    v_p(\bm{\rho}) =
    \begin{cases}
        v_F(\bm{\rho}) &\mbox{if } s(\bm{\rho}) = p\\
        v_R(\bm{\rho}) &\mbox{otherwise} 
    \end{cases}
    \quad \forall p \in \mathcal{P}
\end{equation}
Assuming a well-trained detector and a faithful explanation method, heatmaps for \(v_p\) should closely match the manipulated region.
Since an objective ground-truth is available by construction, it's possible to assess \textit{manipulation detection} using common segmentation metrics.
First, \(M_{\text{in}}\) measures the percentage of heatmap mass inside the ground-truth mask, \ie \(\int_\mathcal{G} m_p(\bm{\rho})h(\bm{\rho}) d\lambda\), to ensure that little or no relevance is assigned to non-manipulated regions.
Second, precision at 100 ($P_{100}$), \ie the fraction of the 100 most relevant pixels that falls inside the ground-truth, accounts for manipulation artefacts significantly smaller than the selected region.
Additional manipulation detection metrics are reported in \cref{app:manipulation-detection-metrics}.

As a point of discussion, both humans and computers may ``look at'' other parts of a video to assess whether one portion is manipulated, \eg noting the mismatch between a smiling mouth and two frowning eyes.
However, when asking ``why is the video fake?'', we expect to be pointed at the visible manipulation and not at other natural-looking features.
Therefore, in this context, manipulation masks are considered as the ground-truth explanation.

\section{Experiments}
\label{sec:experiments}

The previous section establishes a set of desirable qualities of explanations and proposes evaluation metrics built on sound mathematical foundations.
We now consider several techniques from previous works and \emph{quantify} their effect on explanations using these metrics.
\Cref{sec:experiments-results} analyses the effects of
\begin{enumerate*}[label=\roman*),before=\unskip{: },itemjoin={{; }},itemjoin*={{; and }},after={{.}}]
    \item data preparation~\cite{wang_humanunderstandable_2021}
    \item loss-based regularization~\cite{rudin_nonlinear_1992}
    \item augmentation-based regularization~\cite{devries_improved_2017}
    \item model architecture~\cite{geirhos_imagenettrained_2018,tuli_are_2021}
\end{enumerate*}
Both and classification performance~(\cref{tab:dfdc-dfd-classification}) and explanation quality~(\cref{fig:dfdc-dfd-explanation}) are reported for each experiment.
Furthermore, \Cref{sec:experiments-user-study} discusses post-processing techniques for heatmap visualization, which are important for communicating explanations to users in practice.

\paragraph{Training dataset.}
All models are trained on videos from the \acrlong{DFDC}~\cite{dolhansky_deepfake_2020} in ``high-quality'' compression (constant rate quantization 23).
Specifically, we train on $19k$ real and $100k$ fake videos, and use the official validation split of $2k$ real and $2k$ fakes for hyper-parameter tuning.
Each video is preprocessed using the MTCNN face detector~\cite{zhang_joint_2016}, the main face is heuristically determined among all detections, then cropped and resized to $224\!\times\!224$ pixels.
Part segmentation is obtained with the BiSeNet face parser~\cite{yu_bisenet_2021} and aggregated into \textit{background, face, nose, mouth, eyes, ears}.
Additional details on data preparation and dataset statistics are provided in~\Cref{app:datasets}.

\paragraph{Explanation datasets.}
For a cross-dataset evaluation of explanation quality metrics we employ a held-out subset of \acrshort{DFDC}, which has a distribution similar to training videos, and a subset of the \acrfull{DFD}\cite{nickdufour_dfd_2019}, which is more challenging due to the potential distribution shift.
Visual quality metrics (\cref{sec:method-metrics-visual-quality}) are computed on the explanations of fake videos, while manipulation detection (\cref{sec:method-metrics-manipulation-detection}) is evaluated on three part-swaps per video, namely \textit{eyes}, \textit{mouth} and \textit{nose}.
Notably, manipulation detection can only be evaluated on a subset of temporally and spatially aligned videos due to the part-swapping procedure.
Additional details are provided in \Cref{app:datasets}.
While the proposed metrics can be flexibly applied to any dataset of real-fake video pairs, we release the code for preprocessing, training and evaluation on \acrshort{DFDC} and \acrshort{DFD} to encourage comparison and facilitate reproducibility:
\href{https://github.com/baldassarreFe/deepfake-detection}{github.com/baldassarreFe/deepfake-detection}.

\paragraph{Classifier.}
Our baseline model is a 3D CNN trained with no pre-processing, no regularization and no data augmentation except random color augmentations.
Specifically, the backbone feature extractor is an S3D model~\cite{xie_rethinking_2018} pre-trained on Kinetics~400~\cite{kay_kinetics_2017}.
The output of each convolutional block is pooled, concatenated, and fed to a 2-layer MLP classification head.
Such shortcut connections proved beneficial over a sequential model in early experiments, likely due to the multi-scale nature of manipulation artefacts. 
During training, the AdamW optimizer~\cite{loshchilov_adamw_2019} minimizes a cross-entropy loss \(\mathcal{L}_\text{CE}\) based on binary video labels until a validation loss stops improving.
Additional details about hyperparameters and training can be found in \cref{app:training-details}.
For each model variant described below, \Cref{tab:dfdc-dfd-classification} reports the average cross-entropy loss and AUROC over 3 runs on the official test split.
While all models achieve satisfactory results on both datasets, performance drops when generalizing from \acrshort{DFDC} to \acrshort{DFD}.
For this reason, we consider explanation metrics evaluated on DFDC more indicative of explanation quality in the following experiments.

\begin{table}[bp]
\small
\centering
\caption{\textbf{Classification metrics:}
$L_\text{CE}$ is categorical cross-entropy (\(\downarrow\)), $A_\text{ROC}$ is the area under the receiver operating characteristic curve (\(\uparrow\)).
Average values over 3 runs, full results in \Cref{tab:dfd-classification-full,tab:dfdc-classification-full}.
Reported values account for class imbalance as detailed in \Cref{tab:dataset-sizes}.}
\label{tab:dfdc-dfd-classification}
\vspace{.3cm}
\begin{tabular}{lllll}
\hline
                        & \multicolumn{2}{c}{DFDC test}            & \multicolumn{2}{c}{DFD}                  \\
                        & $\mathcal{L}_\text{CE}$ & $A_\text{ROC}$ & $\mathcal{L}_\text{CE}$ & $A_\text{ROC}$ \\ \hline
S3D Baseline            & \textbf{.447}           & \textbf{89.0}  & .694                    & \textbf{80.2}  \\
S3D Bilateral           & .696                    & 54.2           & .746                    & 45.8           \\
S3D Gaussian            & .542                    & 81.8           & .760                    & 66.4           \\
S3D TV Loss             & \textbf{.460}           & \textbf{87.4}  & \textbf{.698}           & 75.8           \\
S3D Cutout              & .481                    & 87.2           & \textbf{.655}           & \textbf{79.6}  \\
MViT                    & \textbf{.430}           & \textbf{96.4}  & \textbf{.513}           & \textbf{90.0}  \\ \hline
\end{tabular}

\end{table}

\subsection{Quantitative results}
\label{sec:experiments-results}

\begin{figure*}[tp]
  \centering
  \begin{subfigure}[b]{\textwidth}
      \centering
      \includegraphics[height=1.8cm]{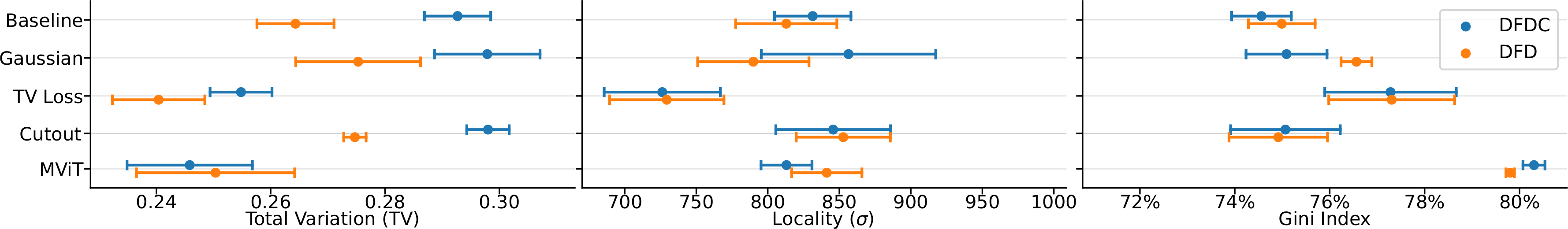}
  \end{subfigure}\vspace{.5cm}
  \begin{subfigure}[b]{\textwidth}
      \centering
      \includegraphics[height=1.8cm]{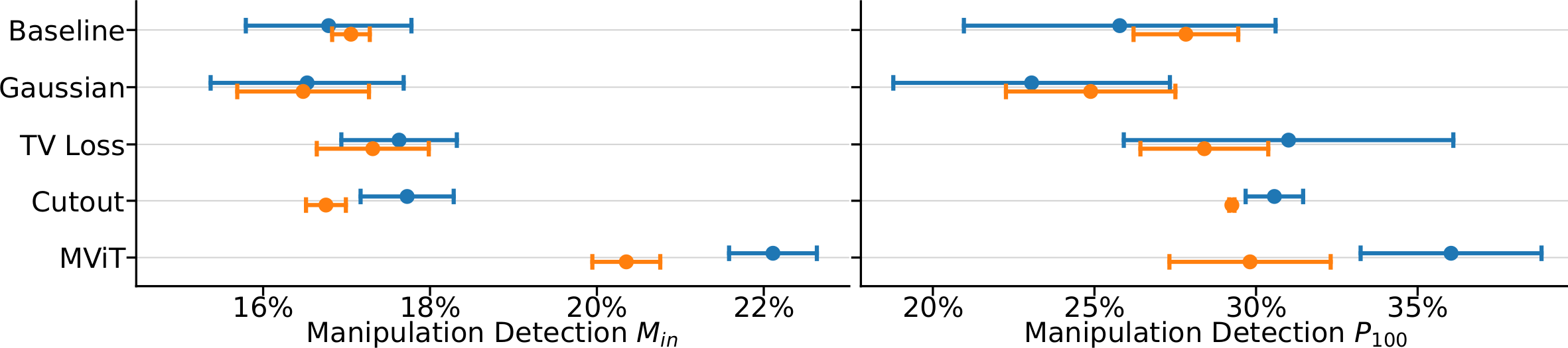}
  \end{subfigure}%
  \caption{\textbf{Quantitative explanation metrics:}
  visual quality (top) and manipulation detection (bottom) for the evaluation subsets of \acrfull{DFDC} and \acrfull{DFD}.
  Higher values indicate better explanation quality, except for \acrshort{TV} and locality $\sigma$.
  Mean and standard deviation of 3 runs, full results in the appendix.}
  \label{fig:dfdc-dfd-explanation}
\end{figure*}

\paragraph{Data preparation.}
As opposed to train-time augmentation, this term indicates transformations that are applied identically to all samples, \eg face detection and cropping described above.
For image \dfs, Wang~\etal observe that generated images have a weaker high-frequency content than real ones~\cite{wang_humanunderstandable_2021} and the explanations of models that rely on this clue are dominated by uninterpretable high-frequency noise.
They suggest pre-processing all samples with a bilateral filter~\cite{tomasi_bilateral_1998} to encourage focussing on other more interpretable features.
In the same spirit, we investigate whether removing high-frequency video components improves smoothness and locality of the explanations in a quantifiable way.

Two variants are considered: a per-frame bilateral filter~\cite{tomasi_bilateral_1998} or a spatio-temporal Gaussian filter; both configured so that common artefacts remain visible.
Only training videos are filtered, leaving validation, test and explanation splits unaltered.
As reported in ~(\cref{tab:dfdc-dfd-classification}), filtered videos result in lower classification performance, which corresponds to the observation in~\cite{wang_humanunderstandable_2021}, and models trained with bilateral filtering fail to converge,thus we exclude them from explanation evaluation.
Disappointingly, blurring does not seem to improve explanation metrics in a consistent way~(\cref{tab:dfdc-dfd-classification}), except for a slightly higher Gini Index that indicates sparser heatmaps.
It is surely possible that stronger filters could produce more marked effects, but at the cost of lower classification performance.
Otherwise, this outcome could be attributed to different generation techniques or compression formats between images and videos.
Nevertheless, we recommend against this type of smoothing preprocessing~\cite{wang_humanunderstandable_2021} for video \dfs until proven more effective.

\paragraph{Regularization loss.}
Regularization refers to training-time techniques that smooth or constrain the loss landscape so that the optimization process yields more desirable solutions that generalize better and/or yield better explanations.
A common technique is to add a per-layer \acrfull{TV} term to the loss function during training~\cite{rudin_nonlinear_1992}.
Considering the activation tensor $\bm{A}^\ell\in\mathbb{R}^{T \times H \times W}$ of an intermediate layer $\ell$, its anisotropic total variation is:
\begin{equation}%
\label{eq:tv-loss}%
    \mathcal{L}^\ell_\text{TV} = \frac{1}{THW} \sum_d \Omega_{\text{1D}}(\bm{A}^\ell_d),
\end{equation}%
where the summation considers all 1D slices of $\bm{A}$ orthogonal to its axes and $\Omega_\text{1D}$ indicates the 1D total variation.
Averaging over all convolutional blocks in our architecture, the optimization objective results $\mathcal{L} = \mathcal{L}_\text{task} + \alpha\mathbb{E}_\ell\left[\mathcal{L}^\ell_\text{TV}\right]$, where $\alpha\in\mathbb{R}^+$ is a hyperparameter.
The additional term places a smoothness constrain on the activations of intermediate layers, which we hope will result in localized peaks in the heatmap corresponding to visible artefacts in the video, though \acrshort{TV} does not control the location of such peaks.

The first effect of \acrshort{TV} regularization is noticeable during the initial phases of training.
For unconstrained models, we observe that \(\exn{\mathcal{L}^\ell_\text{TV}}{\ell}\) tends to increase during the initial phase of training and stabilizes at around $0.5$ after one epoch.
On the other hand, when $\alpha=1$, the optimization process is dominated by $\mathcal{L}_\text{TV}$ for the first epochs and classification loss starts decreasing only after this term drops below $0.1$.
From the results in \Cref{tab:dfdc-dfd-classification}, a strong TV regularization affects classification performance negatively.
However, we also observe a significant improvement over the baseline for locality, sparsity, and manipulation in \Cref{fig:dfdc-dfd-explanation}.
In fact, the average $\sigma$ for DFDC decreases from 814 to 726, indicating more spatially-focused explanations.
Also, the Gini Index increases from 75\% to 77\%, meaning that fewer pixels are responsible for the bulk of the heatmaps.
With respect to manipulation detection, the heatmaps produced by TV-regularized models match more closely the ground-truth, resulting in higher $P_\text{100}$ for both DFDC and DFD.

\paragraph{Video cutout.}
Cutout data augmentation which can greatly improve classification performance by masking input patches at random during training~\cite{devries_improved_2017}.
We adapt Cutout to video data by replacing masking with heavy spatio-temporal blur:
since motion blur occurs naturally, the augmented samples are maintained closer to the data manifold.
We expect Facial Cutout to guide the network towards more meaningful representations, where the relationship between semantic parts of the face are better understood, hence improving part-based manipulation detection.
On the other hand, removing parts of the input might yield more spread out heatmaps, as the network learns to capture information from more diverse locations.
In our experiments, we observe slightly better generalization to \acrshort{DFD} for regularized models~(\Cref{tab:dfdc-dfd-classification}), which confirms the regularization properties of Facial Cutout.
However, the effects on explanations are limited, resulting in slightly higher \acrlong{TV} and manipulation detection scores~(\Cref{fig:dfdc-dfd-explanation}).

\paragraph{Architecture.} The architecture of a model represent a strong inductive biases on what features can be easily learned~\cite{geirhos_imagenettrained_2018,tuli_are_2021}.
As an alternative to the baseline S3D model, we consider another \sota architecture for video classification, namely a multi-scale vision transformer (MViT)~\cite{fan_multiscale_2021}.
The former, based on 3D convolutions, begins with forming local representations which are aggregated into more complex features in later layers.
The latter, based on attention, allows all layers to attend to the input as a whole and encourages representation learning through progressive token aggregation.
We expect the different inductive biases and information flow to affect the explanation heatmaps generated by these architectures.
In particular, we fine-tune the MViT-B $16 \times 4$ variant with the default hyperparameters: random color augmentation, temporal subsampling, cosine learning rate annealing, and weight initialization from Kinetics 400.
For ease of comparison, MViT explanations are obtained with SmoothGrad while attention-specific methods are left to future work.

For the classification task, MViT achieves the best performance on the two datasets~(\cref{tab:dfdc-dfd-classification}), which we attribute to the longer training cycle.
The explanation heatmaps obtained with this architecture are also significantly smoother (TV) and sparser (Gini Index) than CNN-based models, while spatial locality remains similar ($\sigma$).
Furthermore, the bottom row of \cref{fig:dfdc-dfd-explanation} indicates that MViT heatmaps are stronger detectors of manipulated areas, focusing most of the heatmap inside the ground-truth mask ($M_\text{in}$).
We attribute these promising results to
\begin{enumerate*}[label=\roman*),before=\unskip{: },itemjoin={{; }},itemjoin*={{; and }},after={{.}}]
    \item a more robust classifier which can better distinguish fake videos and is thus likely to have learned a good representation of manipulation artefacts
    \item the underlying inductive bias of attention and its effect on gradient propagation used for heatmap generation
\end{enumerate*}

\subsection{Communicating explanations}
\label{sec:experiments-user-study}

As discussed, gradient-based explanations often appear too noisy for users to easily parse.
We propose four simple techniques to post-process heatmaps into increasingly more structured visualizations
\begin{enumerate*}[label=\roman*),before=\unskip{: },itemjoin={{; }},itemjoin*={{; and }},after={{.}}]
\item \textbf{enhanced heatmaps}, clip extreme values and smooth to eliminate high-frequency noise
\item \textbf{gaussian matching}, draw an ellipse corresponding to the mean and variance of each frame
\item \textbf{blob detection},  run a DoG blob detector~\cite{collins_meanshift_2003,pedregosa_scikitlearn_2011} and highlight each blob according to its relevance
\item \textbf{semantic aggregation}, aggregate the heatmap into semantic regions and highlight each part based on its relevance
\end{enumerate*}

\begin{figure*}[tp]
  \centering
  \begin{subfigure}[t]{.64\textwidth}
    \includegraphics[width=\linewidth]{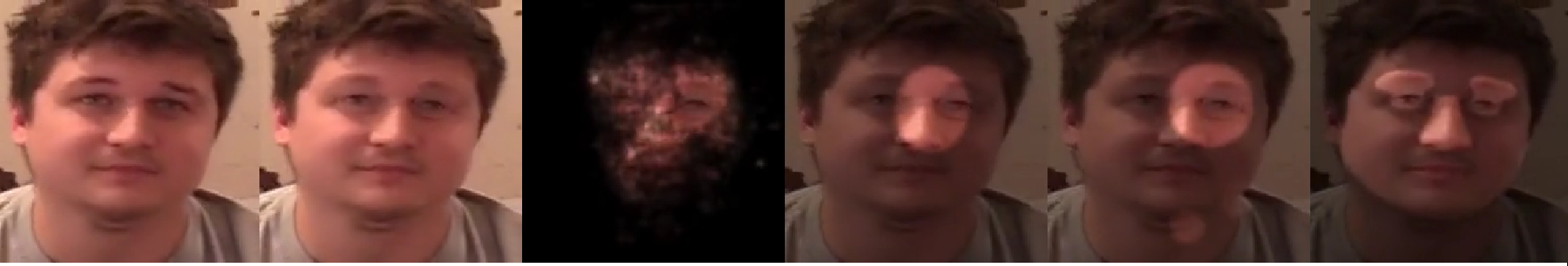}
    \caption{From left to right: real video, fake video, enhanced heatmap, Gaussian matching, blob detection, semantic aggregation.}
    \label{fig:user_study_examples}  
  \end{subfigure}\hfill%
  \begin{subfigure}[t]{.34\textwidth}
    \centering
    \includegraphics[width=\linewidth]{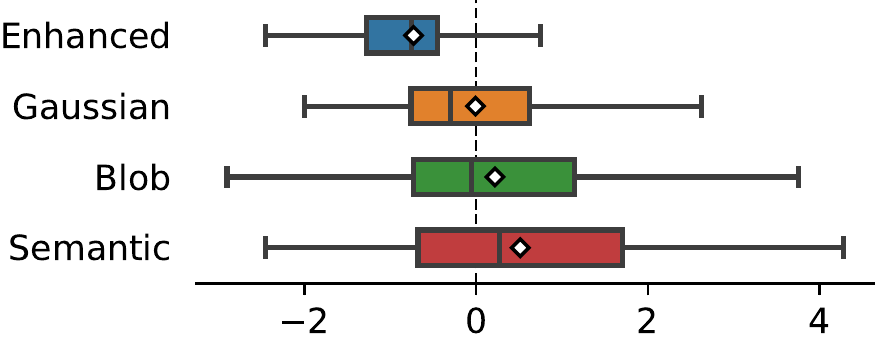}
    \caption{User ratings for alternative explanation visualizations.}
    \label{fig:user-study-results}
  \end{subfigure}
  \vspace*{3mm} %
  \caption{\textbf{User study.}
  \textit{Left panel}: an example of the visualizations submitted to human observation for the study.
  \textit{Right panel}: box plot of normalized score for each explanation visualization technique which  semantic aggregation as preferred.}
\end{figure*}

A small-scale user study (34 participants) is carried out to quantify user satisfaction with respect to each of these visualization techniques.
Each user is presented a set of 10 videos as in \cref{fig:user_study_examples} and is asked to rate the four visualizations, which appear in random order.
A score of 0 means that the visualization is not helpful to detect the \df, while 5 means it easily allowed for its detection.
To minimize appreciation bias, ratings are centered per-user before aggregation by subtracting the average score.
From the results in \cref{fig:user-study-results} we observe a clear relationship between user satisfaction and more structured visualizations.
However, when the classifier performs poorly and heatmaps are uninformative, users will be dissatisfied regardless of post-processing.
However, we also note that when the classifier performs poorly, users will be generally dissatisfied.

\section{Conclusion}
\label{sec:conclusion}

The Explainable AI has developed a plethora of explanation methods of varying degrees of faithfulness.
However, to the best of our knowledge, quantitative metrics to compare the quality of such explanations are lacking.
This work attempts to lay out an objective evaluation framework for \df explanations, which we hope will drive the development of detectors that are better aligned with human cognition.
The main contribution of this paper is the introduction of a family of such metrics, novel or adapted from existing works, to measure visual quality and manipulation detection.

In our experiments we consider several techniques for training \df detectors and study their impact on explainability metrics in a quantitative way, whereas previous work was limited to qualitative comparisons.
We observe that TV regularization has the largest impact across most metrics.
On the other hand, controlling high-frequency components of the input is of little utility, at least when realistic video compression settings are considered.
Finally, we observe that recent architectures such as MViT significantly outperform any of the S3D variations in both detection and explanation quality.
We recommend further study of transformer-based \df classifiers and how to employ attention as an explanation.

\paragraph{Ethical statement.}
As \df technology becomes increasingly accessible, so is the potential for malicious use.
It is therefore urgent to present society with the necessary tools to address this problem and facilitate the safe and ethical use of these creations.
We believe this line of work can bring positive societal impact by facilitating good governance and wider adoption of \df detectors across all media.
Furthermore, more explainable \df detectors can be used to educate the public to better distinguish between real and generated content.
From their perspective, users must feel confident about the technologies that routinely affects their interactions, which may otherwise fall victim to mistrust.

\paragraph{Limitations and future work.}
This project leads to many natural avenues for future research in \acrlong{XAI}.
First, although the proposed metrics are drawn from existing literature and are based on sound mathematical foundations, an extensive study of the correlation between these metrics and human preference would increase their reliability.
Second, it is surely possible to conceive more refined metrics for \df detection to address the shortcomings discussed in \Cref{sec:method-metrics}.
For instance, we have already mentioned that locality (\(\sigma\)) favors unimodal over multimodal heatmaps, whereas more faceted metrics of localization are desirable.
Third, as made evident from the experiments on \dfd, when classification performance is not perfect explanations can be meaningless.
Thus, combining explanations and uncertainty estimation would provide a more complete picture of any \df detector.
Finally, we remark that the proposed metrics are not meant to supplant human judgment, \eg user studies, but rather to provide a non-interactive and repeatable benchmark that is more suitable for guiding the development and facilitating the deployment of better \df detectors.

\bibliography{bibliography}

\clearpage
\appendix
\begin{center}
    {\Large \bf \textcolor{bmvcblue}{Supplementary Material}}
\end{center}

\section{Datasets}
\label{app:datasets}

\subsection{DeepFake Detection Challenge}

The \acrshort{DFDC} dataset was released as part of the homonymous Kaggle challenge~\cite{dolhansky_deepfake_2020}.
It contains approximately 120k videos, the majority of which are \dfs created with different manipulation methods.
The dataset comes in different variants of compression quality, which are mostly relevant for training robust classifiers and assessing detection performance.
For an analysis of explanation quality, we choose to work on high-quality videos where we expect manipulation artifacts to be more prominent.
\Cref{tab:dataset-sizes} reports how many videos are used for training, validation, testing and explanation evaluation.

\paragraph{Preprocessing.}
The videos from the dataset might contain one or more faces.
Of these, only one is manipulated in the case of ``fake'' videos.
For the purpose of training the \dfs classifier, each video is preprocessed as follows:
\begin{enumerate}[noitemsep,nolistsep]
\item Videos are spatially resized with padding so that each frame is \(640\times640\) pixels;
\item MTCNN is applied every 5 frames, which outputs rectangular bounding boxes tightly cropped
      around all faces in a frame
\item Face detections are linked across frames using a greedy overlap-based heuristic;
      namely, if two bounding boxes overlap with $\text{IoU}>0.5$ they are considered the same face;
\item The longest consecutive sequence of linked boxes is considered the main face and is assumed to be the target for \df manipulation,
      all other boxes are discarded and the video is clipped to the frames containing the main face;
\item For intermediate frames where MTCNN was not applied, a bounding box for the main face is created
      by linearly interpolating the corners of the two closest boxes;
\item Boxes are expanded by $1.5\times$ to capture more of the hair, neck and background
\item All frames belonging to the main face sequence are cropped according to their box;
      rectangular crops are resized to $224 \times 224$ and used for training;
\item BiSeNet is applied every 5th frame of a $512 \times 512$ version of the aforementioned video,
      the probabilistic output of BiSeNet is resized with bilinear interpolation to match the original size and then the most likely face part is selected.
\end{enumerate}

\paragraph{Classification.}
For training, validation, and testing, each video is processed separately.
This means that fake videos will not be perfectly aligned with the corresponding real video, neither in space nor in time.
Also, it means that detection and parsing might fail on some fake videos, due to the low quality of the manipulation.
While this can hinder training, it also represents a realistic scenario where unseen videos are submitted to a trained classifier.

\paragraph{Expalantion evaluation.}
Explanation metrics based on manipulation detection require perfectly-aligned pairs of real-fake videos.
Therefore, face detection and parsing are performed on real videos and applied identically to all corresponding fake videos.
Since the pairing between real and manipulated videos is only given for the training split of \dfdc, and only some videos are perfectly aligned, we use an held-out subset of the training split consisting of 230 fake videos created from 100 real videos.

\subsection{DeepFake Detection Dataset}

The DeepFake Detection Dataset~\cite{nickdufour_dfd_2019} is constructed by recording actors in various situations and then applying face-swapping \df techniques to the videos.
The dataset contains a large number of videos, but only a limited number of them are perfectly aligned and can be used for evaluating explanations on part-based manipulation detection.
With respect to the classification task, this dataset is never used during training and it represents a good benchmark for out-of-distribution generalization.
\Cref{tab:dataset-sizes} reports the number of videos used to evaluate classification performance and explanation metrics.
The videos are preprocessed in the same way as \acrshort{DFDC}.
For manipulation detection, perfect alignment is available for 107 fake videos created from 37 real ones.

\begin{table}[h]
\centering
\caption{
\textbf{Dataset sizes:} number of real and fake videos contained in each dataset and split.
DFDC is used to train all classifiers in this work, to report classification metrics, and to evaluate explanation metrics.
DFD is only used for testing and explanation evaluation.
}
\label{tab:dataset-sizes}
\vspace{1cm}
\begin{tabular}{@{}rrrrr@{}}
\toprule
            & \multicolumn{2}{c}{DFDC} & \multicolumn{2}{c}{DFD} \\
            & Real        & Fake       & Real       & Fake       \\ \midrule
Train       & 19143       & 99953      &    -       &    -       \\
Validation  &  1975       &  1968      &    -       &    -        \\
Test        &  2479       &  2486      &   37       &  107       \\
Explanation &   100       &   230      &   37       &  107       \\ \bottomrule
\end{tabular}
\end{table}
\clearpage
\section{Classification performance}
\label{app:classification-results}

We report relevant classification metrics for the test split of DFDC in \Cref{tab:dfdc-classification-full} and for a subset of DFD in \Cref{tab:dfd-classification-full}.
In addition to cross-entropy loss ($\mathcal{L}_\text{CE}$) and area under the receiver operating characteristic curve ($A_\text{ROC}$), we also report precision, recall, and F1 scores obtained when the model output is binarized with a threshold of 0.5. Furthermore, we report average precision (AP), \ie the area under the precision-recall curve as the classification threshold changes.
For each metric and configuration described in \Cref{sec:method-metrics} and \Cref{sec:experiments}, we report mean and standard deviation of 3 runs.

\begin{table*}[ht]
\scriptsize
\centering
\caption{Classification metrics for DFDC, test split. For each configuration and metric, mean and standard deviation of 3 runs are reported. For all metrics except the \(\mathcal{L}_\text{CE}\) loss, values are given in percentage and a higher value indicates a better result.}
\label{tab:dfdc-classification-full}
\vspace{1cm}
\begin{tabular}{lrlrlrlrlrlrl}
\toprule
{} & \multicolumn{2}{c}{$\mathcal{L}_\text{CE} \downarrow$} & \multicolumn{2}{c}{Precision \(\uparrow\)} & \multicolumn{2}{c}{Recall \(\uparrow\)} & \multicolumn{2}{c}{F1 \(\uparrow\)} & \multicolumn{2}{c}{AP \(\uparrow\)} & \multicolumn{2}{c}{$A_\text{ROC} \uparrow$} \\
{}                      &   avg & std   &      avg & std  &   avg & std  &   avg & std  &   avg & std  &   avg & std  \\
\midrule
S3D Baseline            & 0.447 & 0.036 &    80.89 & 1.82 & 82.27 & 4.01 & 81.50 & 1.17 & 88.79 & 2.19 & 89.02 & 1.08 \\
S3D Bilateral           & 0.696 & 0.003 &    54.00 & 0.15 & 32.06 & 8.47 & 39.91 & 6.74 & 52.75 & 0.58 & 54.24 & 0.67 \\
S3D Gaussian            & 0.542 & 0.031 &    78.49 & 1.51 & 64.26 & 2.89 & 70.65 & 2.16 & 80.20 & 3.30 & 81.77 & 2.33 \\
S3D TV Loss             & 0.460 & 0.027 &    78.68 & 1.53 & 81.04 & 3.50 & 79.82 & 2.12 & 88.24 & 1.78 & 87.41 & 1.88 \\
S3D Cutout              & 0.481 & 0.037 &    78.39 & 0.73 & 82.72 & 4.10 & 80.46 & 1.87 & 86.42 & 3.56 & 87.19 & 2.14 \\
MViT                    & 0.430 & 0.004 &    83.64 & 0.28 & 94.30 & 1.46 & 88.65 & 0.57 & 96.59 & 0.32 & 96.38 & 0.38 \\
\bottomrule
\end{tabular}

\end{table*}

\begin{table*}[ht]
\scriptsize
\centering
\caption{Classification metrics for DFD. For each configuration and metric, mean and standard deviation of 3 runs are reported. For all metrics except the \(\mathcal{L}_\text{CE}\) loss, values are given in percentage and a higher value indicates a better result.}
\label{tab:dfd-classification-full}
\vspace{1cm}
\begin{tabular}{lrlrlrlrlrlrl}
\toprule
{} & \multicolumn{2}{c}{$\mathcal{L}_\text{CE} \downarrow$} & \multicolumn{2}{c}{Precision \(\uparrow\)} & \multicolumn{2}{c}{Recall \(\uparrow\)} & \multicolumn{2}{c}{F1 \(\uparrow\)} & \multicolumn{2}{c}{AP \(\uparrow\)} & \multicolumn{2}{c}{$A_\text{ROC} \uparrow$} \\
{}                      &   avg & std   &      avg & std  &   avg & std   &   avg & std  &   avg & std  &   avg & std  \\
\midrule
S3D Baseline            & 0.694 & 0.080 &    72.48 & 5.55 & 61.64 &  2.81 & 66.59 & 3.73 & 82.88 & 2.67 & 80.24 & 2.31 \\
S3D Bilateral           & 0.746 & 0.006 &    42.37 & 0.56 & 26.97 &  2.51 & 32.94 & 2.05 & 43.90 & 0.00 & 45.82 & 0.29 \\
S3D Gaussian            & 0.760 & 0.055 &    60.51 & 1.77 & 49.03 & 12.73 & 53.66 & 8.34 & 66.03 & 2.37 & 66.42 & 0.90 \\
S3D TV Loss             & 0.698 & 0.038 &    65.21 & 2.03 & 66.79 &  7.17 & 65.90 & 4.25 & 77.66 & 3.72 & 75.75 & 3.59 \\
S3D Cutout              & 0.655 & 0.065 &    72.48 & 5.26 & 59.95 &  4.25 & 65.44 & 1.97 & 82.23 & 0.91 & 79.59 & 0.48 \\
MViT                    & 0.513 & 0.015 &    74.75 & 1.24 & 83.41 &  3.50 & 78.83 & 2.12 & 91.73 & 1.70 & 90.04 & 1.79 \\
\bottomrule
\end{tabular}

\end{table*}

\clearpage
\FloatBarrier
\section{Training details}
\label{app:training-details}

\subsection{High-frequencies smoothing}
For models trained with smoothing preprocessing, either Gaussian blur or bilateral filtering are applied.
Gaussian blur is applied at the video level using a spatial standard deviation of 0.8 and a temporal standard deviation of 0.5.
Bilateral filtering is applied per-frame using a spatial standard deviation of 2 and a color range standard deviation of 0.1.
These values are empirically chosen so that the filtered videos remain qualitatively similar to the original ones and that common \df artefacts are still visible.

\subsection{Default video augmentation}
For all models, color augmentations are applied to training videos to improve generalization.
Each video is augmented with probability 0.5.
If the video is augmented, one of the following transformations  is chosen with equal probability:
grayscale conversion,
RGB shifting,
gamma shifting,
contrast limited adaptive histogram equalization,
hue, saturation and value shifting,
brightness and contrast shifting.

\subsection{Cutout}
When cutout is enabled, each video is augmented with cutout with probability 0.5.
Cutout acts on a $64 \times 64$ region of the video selected with uniform probability.
Once a mask is selected, its contents are blurred with a strong Gaussian filter (standard deviation 4).

\subsection{Architectures and training details}

\paragraph{Multiscale S3D.}
The S3D architecture consists of several inception blocks using separable 3D convolution layers.
We use a model pretrained on Kinetics~400 as the backbone feature extractor for the \df classifier.
On top of the original architecture, we add shortcut connections from intermediate layers to the classification head,to allow easier access to multiscale features which might be relevant for the task.
Specifically, we collect the input activations of the 2nd, 3rd, 4th and 5th pooling layers.
These activations are first average-pooled to the size of the smallest one, concatenated, processed through $1 \times 1 \times 1$ convolution, and eventually pooled into a 128-dimensional feature vector.
The classification head is a simple 2-layer MLP with output size of 2 and softmax activation.

During one epoch of training, real videos are sampled more than once to match the number of fake videos in the training set. 
From each video, a clip of 64 consecutive frames is extracted at random.
Videos shorter than 64 frames are padded by appending black frames.
No spatial cropping is performed since the videos already contain centered faces.
The optimizer processes mini-batches of 32 videos at the time.
The learning rate of the Adam optimizer is set to $10^{-4}$ for pretrained parameters and to $10^{-3}$ for the classification head.
An additional weight decay loss is applied to all parameters except biases with strength $10^{-5}$.

For validation, only the first 64 frames of each video are considered and no augmentations are applied.
Training runs for 5 epochs, unless validation loss stops decreasing, in which case early stopping is applied.
The results reported in the tables are relative for videos in the training set.
For these, we average the output probabilities of 5 equally-spaced 64-frames clips from each video.

\paragraph{Multiscale ViT.}
As an alternative to the 3D CNN backbone, we experiment with a transformer architecture.
Specifically, we use a Multiscale Vision Transformer that combines attention layers with multiscale hierarchical processing.
For this model, we maintain the original architecture except for the classification head that is modified to output 2 classes.
Similarly to S3D, the weights are initialized from a model pretrained on Kinetics~400.

Training, validation and testing follow the default settings from the authors.
Namely, the learning rate follows a cosine annealing schedule without warm-up, random temporal crops of 16 frames are selected from each training video, multiple temporal crops are considered for testing, spatial cropping is disabled.

\paragraph{Compute resources.}
All models are trained on a single machine equipped with 4 NVIDIA V100 GPUs with 32GB of RAM each, which allow for large batch sizes.
Once trained, the model can be ran for both inference and explanations on more modest hardware, \eg a single GPU environment with 12GB of RAM.

\clearpage
\FloatBarrier
\section{Explanation metrics}
\label{app:explanation-metrics}

This section details how the explanation metrics introduced in \Cref{sec:method-metrics} are computed in practice using discretized videos, masks and heatmaps.
Furthermore, \Cref{tab:dfdc-explanation-full} and \Cref{tab:dfd-explanation-full} report the metrics for all variations considered in this work as the average and standard deviation of 3 runs each.

The main text uses a compact notation where videos are defined as a mapping from the discretized grid
\(\mathcal{G} = \{1,\ldots,T\}\times\{1,\ldots,H\}\times\{1,\ldots,W\}\)
to RGB pixel values.
For this appendix, we choose a more explicit notation based on tensors.
We remark the equivalence between the two notations since any function \(f: \mathcal{G} \rightarrow R^+\) can be uniquely represented as a \(T\!\times\!H\!\times\!W\) tensor whose element at \((t,h,w)\) is \(f(t,h,w)\).
In this context, it is useful to define the derivative and integral operators \(\nabla\) and \(\int\) as:
\begin{align}
    \left(\nabla f(\bm{\rho})\right)_{i=1,2,3} &=f(\bm{\rho}+\bm{e}_i)-f(\bm{\rho}), \\
    \int_{\mathcal{G}}f d\lambda &= \frac{1}{THW}\sum_{\bm{\rho}\in \mathcal{G}} f(\bm{\rho}),
\end{align}
where the vector \(\bm{\rho} = (t,u,v)\) denotes the pixel coordinates,
and the vectors \(\bm{e}_i\) are the usual orthonormal basis i.e. \((e_i)_j =\delta_{ij}\).

\subsection{Total variation}
\noindent Total variation is used to measure the smoothness of a heatmap and is defined as:
\begin{equation}
    \operatorname{TV}(h) = \frac{1}{THW} \sum_{\bm{\rho}\in\mathcal{G}} \nabla h(\bm{\rho}),
\end{equation}
where the discrete gradient $\nabla h$ at coordinates \(\bm{\rho}=(t,u,v)\) is computed as:
\begin{equation}
    \left| h(t,u,v)- h(t+1,u,v)\right| +
    \left| h(t,u,v)- h(t,u+1,v)\right| +
    \left| h(t,u,v)- h(t,u,v+1)\right|
\end{equation}

\subsection{Variance volume}
\noindent To measure the spatial localization of the heatmap, we first compute its mean and variance:
\begin{align}
    \bm{\mu}    &= \sum_{\bm{\rho}\in\mathcal{G}} \bm{\rho} h(\bm{\rho}), \\
    \bm{\Sigma} &= \sum_{\bm{\rho}\in\mathcal{G}} (\bm{\rho}-\bm{\mu})(\bm{\rho}-\bm{\mu})^T h(\bm{\rho}),
\end{align}
where the vector $\bm{\rho}=(t,u,v)^T$ represents pixel coordinates.
Then, to summarize the $3 \times 3$ variance matrix as a scalar, we consider its volume given by the determinant $|\det(\bm{\Sigma})|$.
Larger volumes correspond to more spread out heatmaps, while smaller values indicate more localized explanations.
Importantly, this metric is particularly indicated for unimodal heatmaps that focus around a single location of the video.

\subsection{Gini Index}
\noindent The Gini Index was initially introduced as an economic indicator of wealth distribution~\cite{gini_variabilita_1912}, but it is considered a good measure of sparsity due to its properties~\cite{hurley_sparsity_2009}.
The Gini Index of an heatmap measures is defined as:
\begin{equation}
    \operatorname{G} = 
    \frac{2}{THW} 
    \frac{\sum_i i \cdot h(\bm{\rho_i})}{\sum_i h(\bm{\rho_i})}
    - \frac{THW+1}{THW},
\end{equation}
with the indices $i=1,\ldots,THW$ that select pixel coordinates such that $h(\bm{\rho}_i) \leq h(\bm{\rho}_{i+1})$.
Notably, the ``sparsity'' measured by the Gini Index refers to the scalar importance values of each pixel and not their location in the heatmap.
The heatmap will have a high Gini Index if most of the explanation mass is concentrated in few highly-relevant pixels while all other pixels have low relevance.

\subsection{Faithfulness}
\label{app:deletion-scores}
Faithfulness is generally used to compare explanation methods on the basis of how closely they identify portions of the input that are meaningful for the classifier and a particular decision.
Faithfulness is measured using the \emph{deletion score}, which represents the area under the curve traced by the confidence in $p(\textsc{fake}|v)$ as pixels are removed from the video in decreasing order of relevance.
Considering the large amount of pixels in a video, the curve is approximated by removing several pixels in one step.
Specifically, we consider the sorted values of an heatmap $h$ and group them in 25 bins.
These bins do not necessarily contain the same amount of pixels, but the total relevance in each bin is approximately the same.

In this work, we are interested in the properties of explanations rather than of explanation methods.
However, a faithful explanation method is a prerequisite for further evaluation of explanation quality.
As a preliminary step, we compare the four explanation methods discussed in \cref{sec:method-explanations} and choose the most faithful one based on its deletion scores on the baseline model.
The following hyperparameters are used:
in SmoothGrad, gradients are averaged over 25 randomly perturbed videos with a noise parameter equal to 0.15 of the RGB range;
in Integrated Gradients the path integral is calculated \wrt a black video baseline using 25 interpolation steps.

With respect to fake videos in DFDC and DFD, the four methods achieve the following average deletion scores: Sensitivity 42.54\%, Gradient$\times$Input 43.68\%, SmoothGrad 41.25\%, Integrated Gradients 43.77\%.
Therefore, SmoothGrad has the lowest deletion score of the four (p value of paired one-sided t-test \(<10^{-5}\)) and is then employed throughout all experiments.

\subsection{Manipulation detection}
\label{app:manipulation-detection-metrics}
The ability to detect and localize manipulations is measured by considering videos created by overlaying a portion of a fake video $v_F$ to its corresponding original video $v_R$.
The blending mask is obtained by selecting a face part $p\in\{\text{mouth},\text{nose},\text{eyes}\}$ from those extracted with BiSeNet~\cite{yu_bisenet_2021}.
The explanation heatmap can then be compared with the ground-truth manipulation mask to determine whether the model is focusing on manipulated regions of the video.
Without loss of generality, only the first 64 frames of each video are considered in all manipulation detection evaluations.

\begin{figure}[ht]
\centering
\begin{subfigure}[b]{0.45\textwidth}
    \centering
    \includegraphics[width=\textwidth]{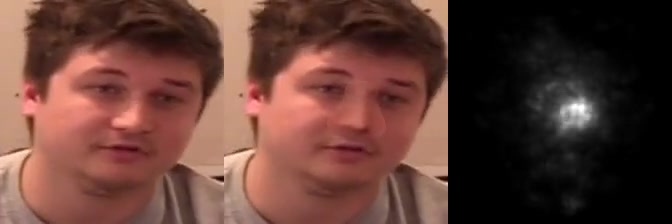}
\end{subfigure}\hfill
\begin{subfigure}[b]{0.45\textwidth}
    \centering
    \includegraphics[width=\textwidth]{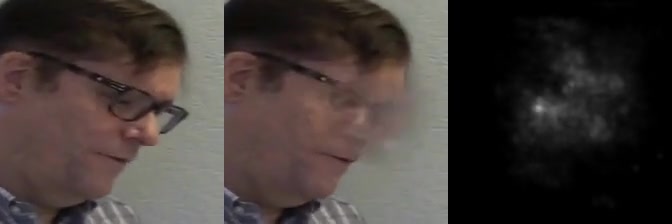}
\end{subfigure}
\begin{subfigure}[b]{0.45\textwidth}
    \centering
    \includegraphics[width=\textwidth]{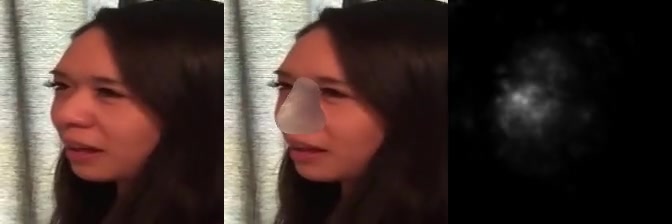}
\end{subfigure}\hfill
\begin{subfigure}[b]{0.45\textwidth}
    \centering
    \includegraphics[width=\textwidth]{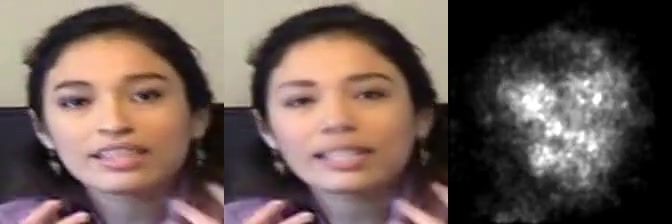}
\end{subfigure}
\begin{subfigure}[b]{0.45\textwidth}
    \centering
    \includegraphics[width=\textwidth]{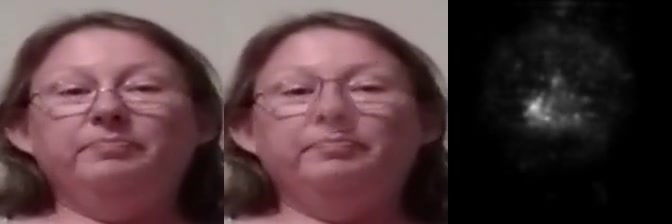}
\end{subfigure}\hfill
\begin{subfigure}[b]{0.45\textwidth}
    \centering
    \includegraphics[width=\textwidth]{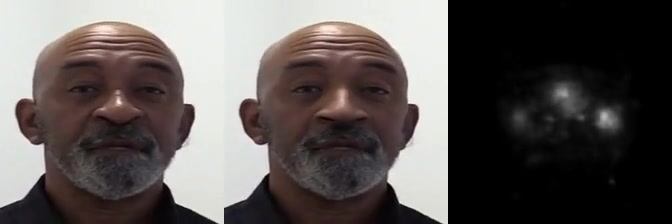}
\end{subfigure}
\begin{subfigure}[b]{0.45\textwidth}
    \centering
    \includegraphics[width=\textwidth]{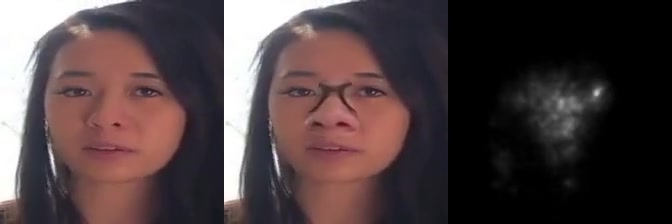}
\end{subfigure}\hfill
\begin{subfigure}[b]{0.45\textwidth}
    \centering
    \includegraphics[width=\textwidth]{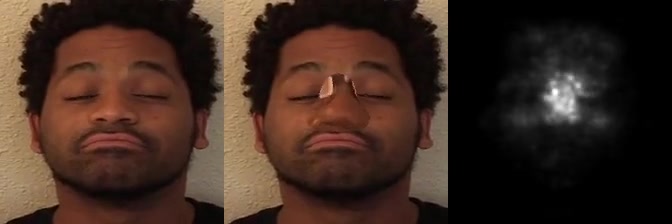}
\end{subfigure}
\begin{subfigure}[b]{0.45\textwidth}
    \centering
    \includegraphics[width=\textwidth]{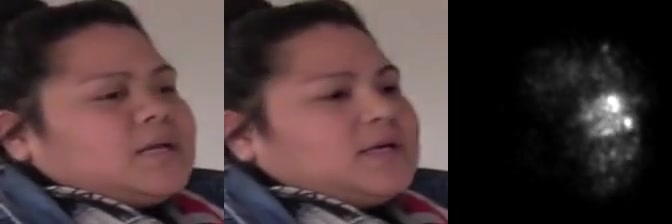}
\end{subfigure}\hfill
\begin{subfigure}[b]{0.45\textwidth}
    \centering
    \includegraphics[width=\textwidth]{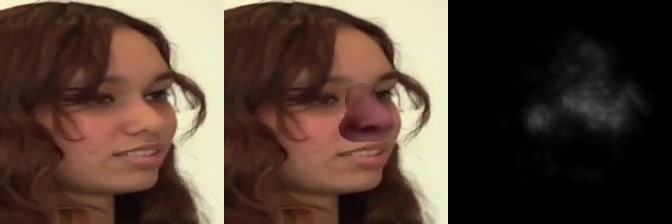}
\end{subfigure}
\begin{subfigure}[b]{0.45\textwidth}
    \centering
    \includegraphics[width=\textwidth]{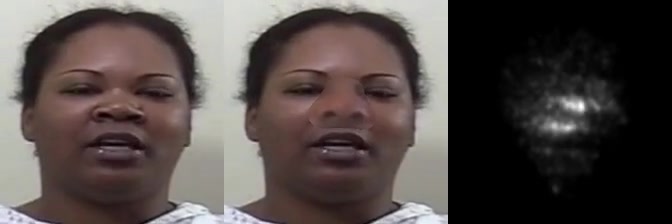}
\end{subfigure}\hfill
\begin{subfigure}[b]{0.45\textwidth}
    \centering
    \includegraphics[width=\textwidth]{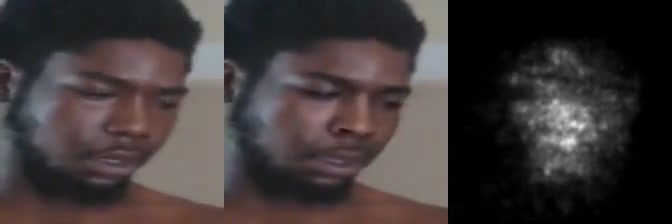}
\end{subfigure}
\begin{subfigure}[b]{0.45\textwidth}
    \centering
    \includegraphics[width=\textwidth]{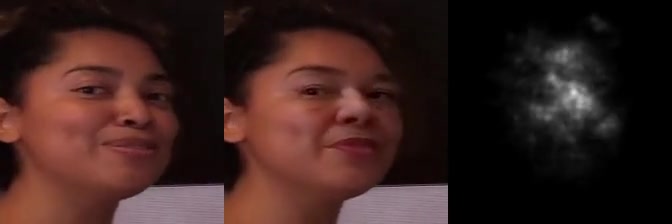}
\end{subfigure}\hfill
\begin{subfigure}[b]{0.45\textwidth}
    \centering
    \includegraphics[width=\textwidth]{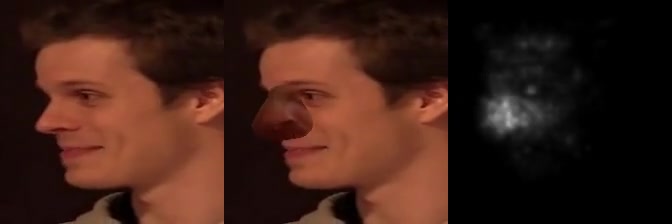}
\end{subfigure}
\caption{Additional examples for manipulation detection. Random frames from random videos. From left to right: original, part-based manipulated video, heatmap.}
\label{fig:additional-examples}
\end{figure}

For the main metrics, we measure the precision of 100 most-relevant pixels in the heatmap, and the percentage of heatmap contained in the ground-truth manipulation mask.
This approach has the advantage of being threshold independent, as opposed to binarizing the heatmap using an arbitrary threshold and computing metrics such as Intersection over Union.
Notably, both metrics penalize explanations that focus outside the ground-truth mask, but can not distinguish whether the heatmap clusters around an actual manipulation artefact or is uniformly scattered inside the mask~(\Cref{fig:additional-examples}).

\begin{table*}[ht]
\scriptsize
\centering
\caption{Explanation metrics for \acrfull{DFDC}. Mean and standard deviation of 3 independent classifiers whose decisions are explained using SmoothGrad.}
\label{tab:dfdc-explanation-full}
\vspace{1cm}
\begin{tabular}{lrlrlrlrlrl}
\toprule
{} & \multicolumn{2}{c}{TV \(\downarrow\)} & \multicolumn{2}{c}{$\sqrt[3]{\sigma} \downarrow$} & \multicolumn{2}{c}{Gini \(\uparrow\)} & \multicolumn{2}{c}{$M_\text{IN} \uparrow$} & \multicolumn{2}{c}{$P_{100} \uparrow$} \\
{} &  avg &   std &    avg &   std &   avg &   std &           avg &   std &           avg &   std \\
\midrule
S3D Baseline            & 0.285 & 0.006 &   814.4 &  31.0 & 74.93 & 1.00 &         17.43 & 0.38 &         29.43 & 1.34 \\
S3D Bilateral           & 0.428 & 0.044 &  1372.6 & 268.8 & 67.89 & 3.50 &         12.15 & 3.32 &         13.47 & 8.77 \\
S3D Gaussian            & 0.292 & 0.004 &   841.4 &  15.1 & 75.55 & 0.44 &         17.38 & 0.51 &         26.31 & 1.19 \\
S3D TV Loss             & 0.256 & 0.003 &   726.3 &  54.8 & 77.40 & 1.97 &         17.68 & 1.00 &         34.46 & 1.85 \\
S3D Cutout              & 0.296 & 0.005 &   839.0 &  40.2 & 75.26 & 1.40 &         17.76 & 0.63 &         30.48 & 1.10 \\
MViT                    & 0.246 & 0.013 &   808.3 &  17.1 & 80.42 & 0.36 &         22.11 & 0.64 &         36.03 & 3.43 \\
\bottomrule
\end{tabular}

\end{table*}
\begin{table*}[ht]
\scriptsize
\centering
\caption{Explanation metrics for \acrfull{DFD}. Mean and standard deviation of 3 independent classifiers whose decisions are explained using SmoothGrad.}
\label{tab:dfd-explanation-full}
\vspace{1cm}
\begin{tabular}{lrlrlrlrlrl}
\toprule
{} & \multicolumn{2}{c}{TV \(\downarrow\)} & \multicolumn{2}{c}{$\sqrt[3]{\sigma} \downarrow$} & \multicolumn{2}{c}{Gini \(\uparrow\)} & \multicolumn{2}{c}{$M_\text{IN} \uparrow$} & \multicolumn{2}{c}{$P_{100} \uparrow$} \\
{} &  avg &   std &    avg &   std &   avg &   std &           avg &   std &           avg &    std \\
\midrule
S3D Baseline            & 0.261 & 0.009 &   827.6 &  34.4 & 74.84 & 0.74 &         17.09 & 0.15 &         28.62 &  0.80 \\
S3D Bilateral           & 0.413 & 0.061 &  1426.3 & 346.3 & 68.92 & 4.95 &         13.47 & 4.25 &         16.10 & 16.48 \\
S3D Gaussian            & 0.273 & 0.011 &   799.2 &  11.0 & 76.54 & 0.17 &         16.83 & 0.82 &         26.87 &  1.16 \\
S3D TV Loss             & 0.244 & 0.005 &   742.1 &  45.3 & 77.08 & 1.70 &         17.20 & 0.92 &         30.09 &  2.72 \\
S3D Cutout              & 0.274 & 0.004 &   838.7 &  46.5 & 75.33 & 1.47 &         16.87 & 0.31 &         29.22 &  0.96 \\
MViT                    & 0.250 & 0.016 &   834.5 &  27.3 & 80.02 & 0.10 &         20.35 & 0.50 &         29.81 &  3.06 \\
\bottomrule
\end{tabular}

\end{table*}

\clearpage
\FloatBarrier
\section{Additional examples}
\label{app:additional-examples}

Additional examples of semantic parsing, manipulation detection (\Cref{sec:method-metrics-manipulation-detection}), and explanation post-processing for the user study (\Cref{sec:experiments-user-study}) are shown below.

\begin{figure}[ht]
\centering
\includegraphics[width=\linewidth]{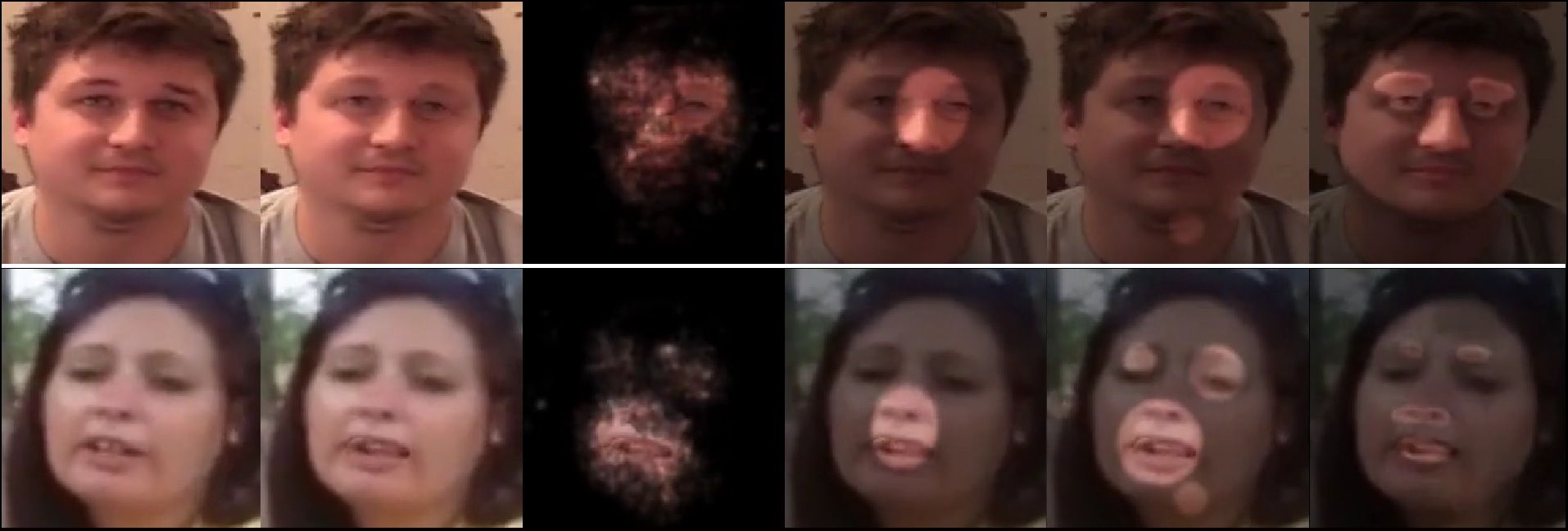}
\caption{User study visualization: real video, fake video, enhanced {heat-map}, Gaussian matching, blob detection, semantic aggregation.}
\end{figure}

\begin{figure}[ht]
\centering
\includegraphics[width=\linewidth]{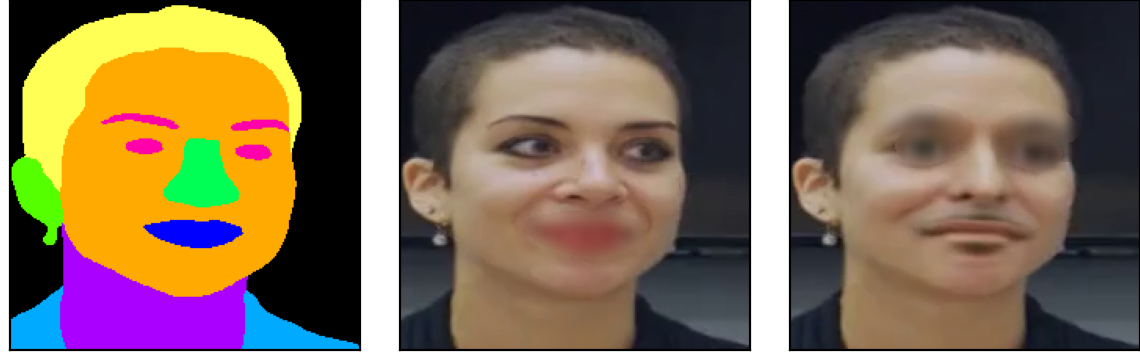}
\caption{Example of semantic parsing as performed by BiSeNet~\cite{yu_bisenet_2021} and of an alternative version of video cutout (\Cref{sec:method-metrics-manipulation-detection}) where heavy blurring is applied dynamically to a semantic region instead of a fixed square.}
\end{figure}

\end{document}